\documentclass[conference]{IEEEtran}
\IEEEoverridecommandlockouts

\usepackage{cite}
\usepackage{amsmath,amssymb,amsfonts}
\usepackage{algorithmic}
\usepackage{graphicx}
\usepackage{textcomp, multirow}
\usepackage{xcolor}
\def\BibTeX{{\rm B\kern-.05em{\sc i\kern-.025em b}\kern-.08em
    T\kern-.1667em\lower.7ex\hbox{E}\kern-.125emX}}
\begin{document}

\title{Dataset Distillation for Quantum Neural Networks
}

\author{\IEEEauthorblockN{Koustubh Phalak}
\IEEEauthorblockA{\textit{CSE Department} \\
\textit{Pennsylvania State University}\\
State College, PA\\
krp5448@psu.edu}
\and
\IEEEauthorblockN{Junde Li}
\IEEEauthorblockA{\textit{Cadence Design Systems}\\
San Jose, CA \\
junde@cadence.com}
\and
\IEEEauthorblockN{Swaroop Ghosh}
\IEEEauthorblockA{\textit{School of EECS} \\
\textit{Pennsylvania State University}\\
State College, PA \\
szg212@psu.edu}
}

\maketitle

\begin{abstract}
Training Quantum Neural Networks (QNNs) on large amount of classical data can be both time consuming as well as expensive. Higher amount of training data would require higher number of gradient descent steps to reach convergence. This, in turn would imply that the QNN will require higher number of quantum executions, thereby driving up its overall execution cost. 
In this work, we propose performing the dataset distillation process for QNNs, where we use a novel quantum variant of classical LeNet model containing residual connection and trainable Hermitian observable in the Parametric Quantum Circuit (PQC) of the QNN. 
This approach yields highly informative yet small number of training data at similar performance as the original data. We perform distillation for MNIST and Cifar-10 datasets, and on comparison with classical models observe that both the datasets yield reasonably similar post-inferencing accuracy on quantum LeNet ($91.9\%$ MNIST, $50.3\%$ Cifar-10) compared to classical LeNet ($94\%$ MNIST, $54\%$ Cifar-10). We also introduce a non-trainable Hermitian for ensuring stability in the distillation process and note marginal reduction of up to $1.8\%$ ($1.3\%$) for MNIST (Cifar-10) dataset.
\end{abstract}


\section{Introduction}
Quantum neural networks (QNNs) have emerged as promising candidates for achieving computational advantages over classical neural networks (NNs) in tasks involving high-dimensional data and noisy training environments \cite{abbas2021power, stein2023benchmarking}. For instance, QNNs exhibit superior scalability for complex problems, requiring $56\%$ fewer parameters and $73\%$ fewer FLOPs compared to classical NNs as problem complexity increases \cite{kashif2024computational}. However, training QNNs on large classical datasets remains prohibitively expensive, with costs scaling linearly with qubit counts (greater than $\$100$K per qubit) \cite{Graps2024, kordzanganeh2023benchmarking}. Each gradient descent step requires numerous quantum executions, making convergence slow and resource-intensive for datasets like MNIST or Cifar-10. 

\begin{figure}[t]
    \centering
    \includegraphics[width=\linewidth]{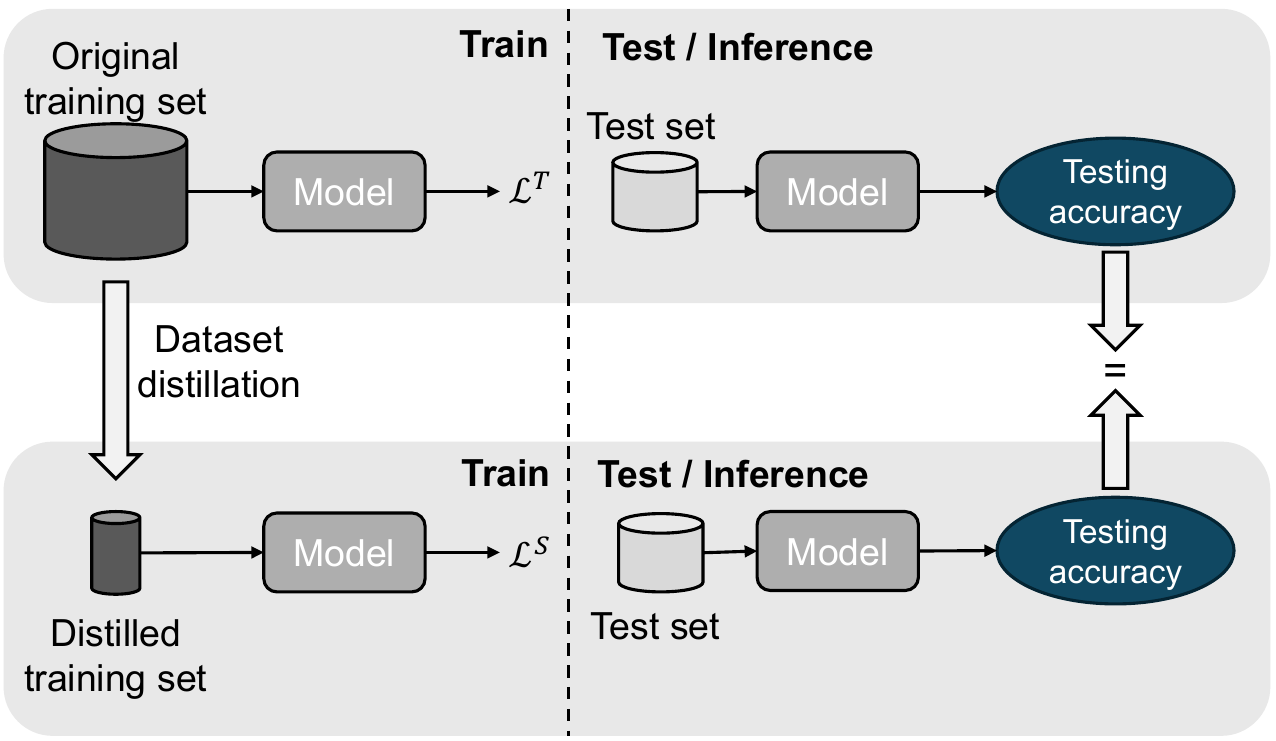}
    \caption{Performance matching dataset distillation process. The goal here is to distill large original training dataset into much smaller synthetic dataset such that when the same model is trained on each of the two datasets in different scenarios, the inferencing performance on the common testing set is nearly the same.}
    \label{fig:performance_matching_dd}
\end{figure}

This challenge motivates the usage of dataset distillation, a technique used to compress large classical dataset into shorter, synthetic data containing all information from the original dataset \cite{wang2018dataset}. \textit{To the best of our knowledge, no prior work on QNNs employ the dataset distillation process to solve this problem.} The results from the original work show that for a model (like LeNet \cite{lecun1998gradient}) having fixed initialization, distilling $60,000$ training samples of MNIST dataset to just 10 synthetic samples (1 image per class) followed by training the model on this distilled data yields close to $94\%$ inferencing accuracy, which is close to $99\%$ from the original training set. Following on these lines, we propose distilling data for QNNs. Specifically, we propose distilling MNIST and Cifar-10 datasets on a novel variant of hybrid quantum-classical LeNet model containing classical feature extraction using convolution layers followed by classical and quantum linear layers for performing classification on the extracted features. Additionally, we add residual connection to the quantum layer to reduce the effect of vanishing gradients, and also use a trainable Hermitian observable while performing measurement in the quantum layer. These modifications help the quantum LeNet to achieve similar accuracy as the classical LeNet in the original work. To ensure stability in the distillation process, we also introduce a non-trainable Hermitian. Our results show that 1. post-distillation inferencing accuracy of quantum LeNet ($91.9\%$ MNIST, $50.3\%$ Cifar-10) is comparable to classical LeNet ($94\%$ MNIST, $54\%$ Cifar-10) and 2. quantum LeNet with non-trainable Hermitian performs marginally worse compared to quantum LeNet with trainable Hermitian ($1.8\%$ less for MNIST, $1.3\%$ less for Cifar-10) with the potential advantage of stabilizing the distillation process.

\begin{figure*}[t]
    \centering
    \includegraphics[width=\linewidth]{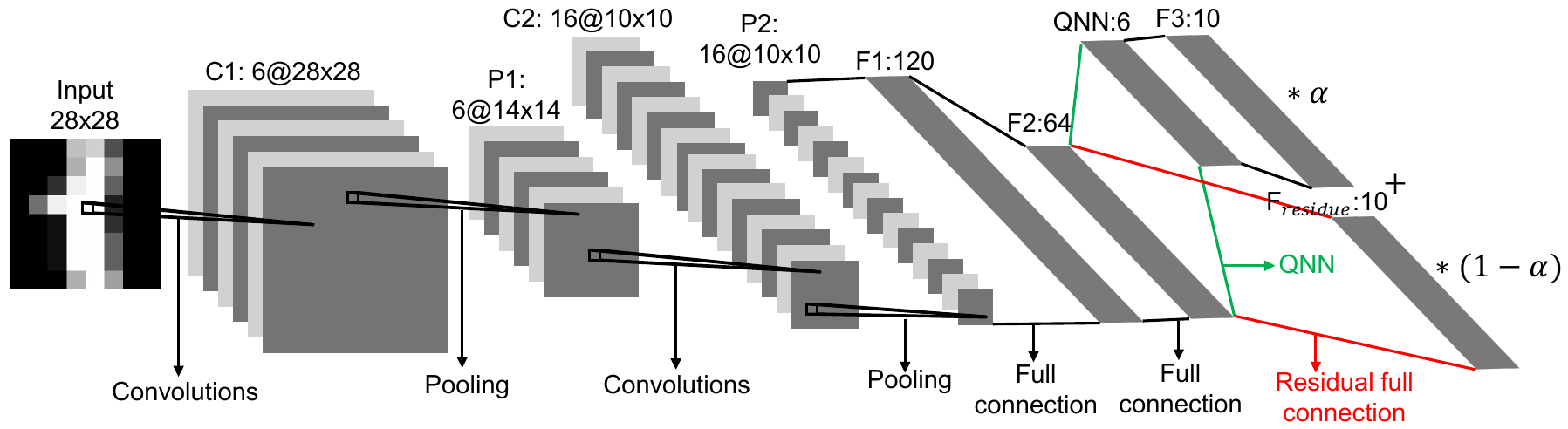}
    \caption{Proposed novel quantum variant of classical LeNet architecture. The convolution layers for feature extraction are kept the same but QNN layer along with residual connection is added in the classifier part between the fully connected dense layers.}
    \label{fig:quantum_lenet}
\end{figure*}

\begin{figure}[t]
    \centering
    \includegraphics[width=\linewidth]{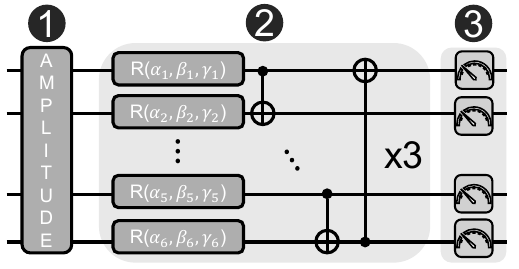}
    \caption{QNN layer architecture containing 1. amplitude embedding, 2. PQC having strongly entangling layers \cite{schuld2020circuit} and 3. measurement with Hermitian observable.}
    \label{fig:qnn_arch}
\end{figure}

The structure of the paper is as follows: Section 2 introduces the relevant background and related works. Section 3 focuses on the modification of the LeNet to its proposed quantum variant. Section 4 shows the results of the distillation process. Finally, we provide discussion in Section 5 and conclude in Section 6.

\section{Background \& Related Works}
\textbf{Quantum Computing:} Similar to classical computing, quantum computing is a paradigm that works on the principles of quantum mechanics \cite{kanamori2020quantum}. The fundamental unit of a quantum computer is a quantum bit (abbreviated as qubit) which is quantum analogue of classical bit. Unlike a bit however, which can be only 0 or 1 at a time, a qubit is represented as a probabilistic state. Specifically, a qubit is represented using the vector $\left[ \begin{smallmatrix} \alpha \\ \beta \end{smallmatrix} \right]$, where $|\alpha|^2$ and $|\beta|^2$ represent the probabilities of qubit value being $0$ and $1$ respectively. This state can be changed with the help of quantum gates, which are unitary matrix operations (i.e. for a matrix $U$, $U^\dagger=U^{-1}$). Quantum gates act on either single qubit (e.g., Hadamard, NOT gate) or on multiple qubits (eg. CNOT, Toffoli gate). An ordered sequence of quantum gates forms a quantum circuit, which can represent various kinds of problems to be solved. A special kind of quantum circuit is Parametric Quantum Circuit (PQC), which contains rotation gates containing rotation value parameters. In Quantum Machine Learning (QML) models, these rotation parameters are often trainable parameters of the PQC.

\begin{figure*}
    \centering
    \includegraphics[width=\linewidth]{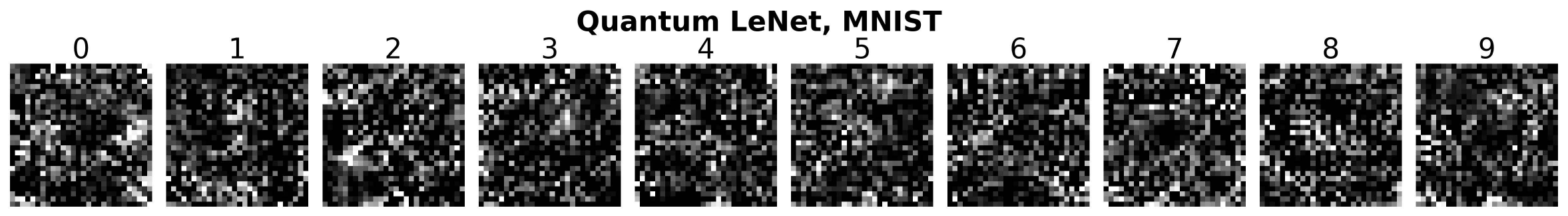}
    \includegraphics[width=\linewidth]{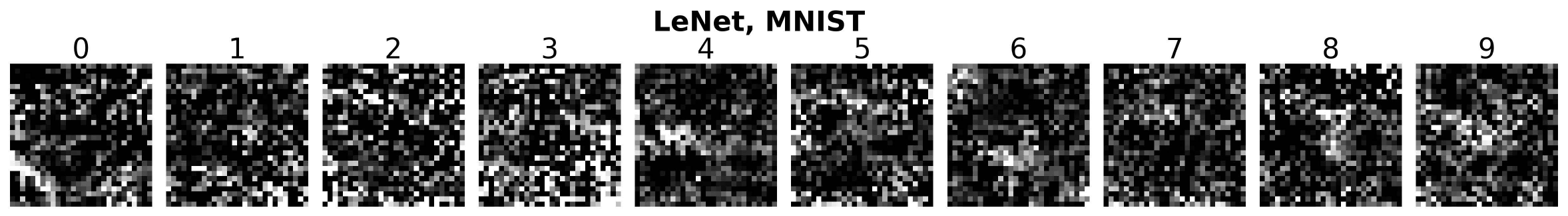}
    \includegraphics[width=\linewidth]{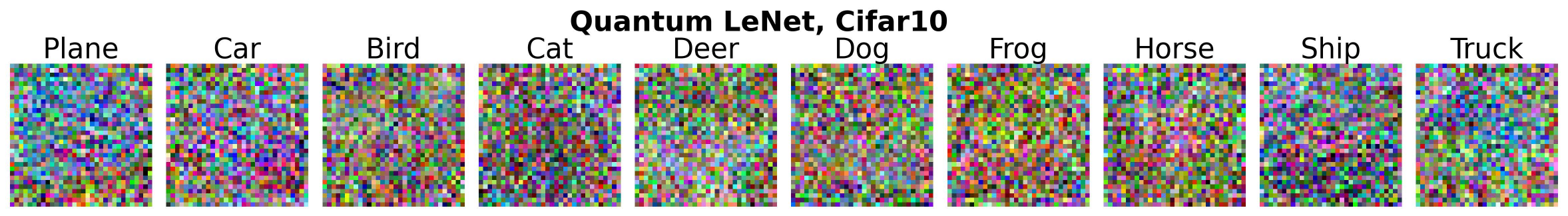}
    \includegraphics[width=\linewidth]{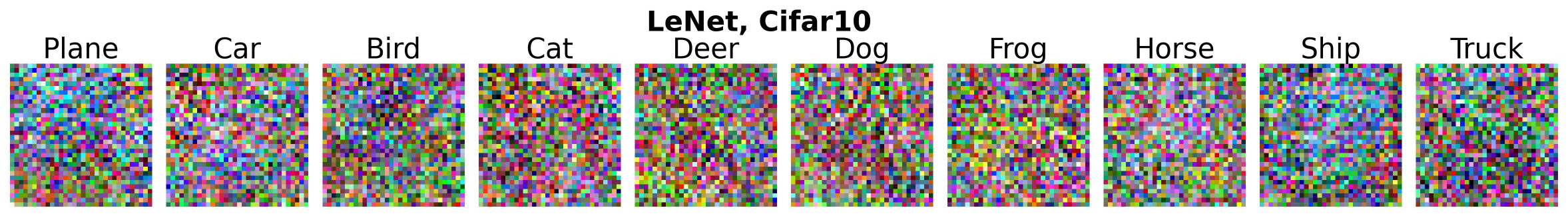}
    \caption{Distilled MNIST and Cifar-10 images on LeNet and Quantum LeNet models.}
    \label{fig:distilled_images}
\end{figure*}

\textbf{Quantum Neural Networks:} QNNs typically consist of three parts: 1. state preparation or embedding circuit that embeds classical data into the quantum Hilbert space, 2. PQC consisting of trainable parameters of QNN and 3. measurement operation in a computational basis (typically Pauli-Z basis) to perform classical measurement of quantum state \cite{nielsen2010quantum}. Following the classical measurement that computes the desired property, loss function computes the loss between the measurement output and expected output which is then used to classically compute gradients and update the PQC rotation parameters.

\textbf{LeNet architecture:} LeNet \cite{lecun1998gradient} is a Convolutional Neural Network (CNN) \cite{li2021survey} that aimed to classify the MNIST dataset. The LeNet model consists of two parts, the first part consisted of sequence of convolution layers that aimed to extract key features of the MNIST digit images, which is followed by a sequence of linear/dense layers that were used to classify the extracted features into the correct class of the image. In this work, we primarily use the LeNet model to base our results. 

\textbf{Dataset distillation:} Unlike classical gradient descent optimization which is a unilevel optimization, dataset distillation is a bilevel optimization process. First, for original training dataset $x$ of size $M$, synthetic dataset $\tilde{x} $ of size $N$ ($M>>N$) is initialized using random noise (such as Gaussian noise) along with a trainable learning rate $\tilde{\eta}$. Then, assuming initial model parameters $\theta_0$, for $T$ iterations and fixed step size $\alpha$ the following steps are performed: 1. One gradient descent step is computed using the synthetic dataset and model parameters are updated ($\theta_1 = \theta_0 = \tilde{\eta}\nabla_{\theta_0}\ell(\tilde{x}, \theta_0)$, where $\ell(\tilde{x}, \theta_0)$ is loss computed on the distilled data using model with initial parameters $\theta_0$), 2. model with updated parameters is used to compute loss on original training step ($\mathcal{L} = \ell(x, \theta_1)$) and finally 3. loss computed in step 2 is used to update the synthetic data and trainable learning rate from the bilevel gradient computation ($\tilde{x} \leftarrow \tilde{x} - \alpha \nabla_{\tilde{x}}\mathcal{L}$,  $\tilde{\eta} \leftarrow \tilde{\eta} - \alpha \nabla_{\tilde{\eta}}\mathcal{L}$).

This method of dataset distillation proposed in the original work \cite{wang2018dataset} is referred to \textit{performance matching} (shown in Fig. \ref{fig:performance_matching_dd}), where the goal is to match the performance of the distilled data on the test set to as close to the original training data. Later, distillation methods with different objectives were proposed in literature such as 1. \textit{parameter matching} \cite{cazenavette2022dataset} that matches the parameter trajectories of models trained on real data and distilled on synthetic data to align model weights as close as possible. This method by further mathematical derivation often reduces to \textit{gradient matching} \cite{zhao2020dataset} that helps in avoiding local minima; 2. \textit{distribution matching} \cite{zhao2023dataset} where the objective is to minimize the distance between the distributions of original training data and distilled data using metrics such as Maximum Mean Discrepancy (MMD). In this work, we focus only on the peformance matching method of dataset distillation for the proposed QNN model.

\begin{table}[t]
\centering
\caption{Post-distillation inferencing accuracy for classical LeNet and different variants of quantum LeNet having fixed initialization. Note that R(NR)=(No) Residual connection, H(NH)=(No) Hermitian observable}
\label{tab:fixed_init_results}
\begin{tabular}{|l|l|lll|}
\hline
\multirow{2}{*}{\textbf{Dataset}} & \multirow{2}{*}{\textbf{\begin{tabular}[c]{@{}l@{}}Classical\\ LeNet\end{tabular}}} & \multicolumn{3}{l|}{\textbf{Quantum LeNet}}                                                \\ \cline{3-5} 
                                  &                                                                                     & \multicolumn{1}{l|}{\textbf{NR, NH}} & \multicolumn{1}{l|}{\textbf{NR, H}} & \textbf{R, H} \\ \hline
\textit{MNIST}                    & 94\%                                                                                & \multicolumn{1}{l|}{82.3\%}          & \multicolumn{1}{l|}{86.5\%}         & 91.9\%        \\ \hline
\textit{Cifar-10}                 & 54\%                                                                                & \multicolumn{1}{l|}{28.5\%}             & \multicolumn{1}{l|}{31.3\%}            & 50.3\%        \\ \hline
\end{tabular}
\vspace{-3mm}
\end{table}

\section{Quantum LeNet Model \& Distillation Setup}
Classical LeNet consists of two parts: the first part extracts features using back-to-back convolution layers and the second part classifies the extracted features using multiple dense layers. This architecture was originally used to classify handwritten MNIST digits in \cite{lecun1998gradient}. For our work, we keep the feature extraction part the exact same as original LeNet architecture and add a $6$ qubit QNN layer in the classifier part between the linear layers that contains, 1. amplitude embedding to embed features and prepare quantum state 2. PQC with trainable parameters (ansatz selected is strongly entangling layers \cite{schuld2020circuit}) and 3. trainable Hermitian measurement observable. Generally, for the observable, a unitary operator such as Pauli-Z operator is chosen that provide measurement value between -1 and 1. However in a deep neural network such a limit on output can cause vanishing gradients \cite{letcher2024tight}. To overcome this restriction, we replace the unitary operator with a trainable Hermitian operator which unlike the unitary operator does not restrict output bounds. To further mitigate the effect of vanishing gradients, we also add residual connections \cite{he2016deep} around the QNN layer. We show the quantum LeNet architecture in Fig. \ref{fig:quantum_lenet}. Additionally, we show the QNN architecture in Fig. \ref{fig:qnn_arch}. 

We use similar experimental setup as the original work \cite{wang2018dataset} where we run the distillation process for one of the two scenarios, where the model initialization is fixed. This is done for both MNIST and Cifar-10 datasets, where MNIST (Cifar-10) dataset is distilled to 10 (100) images with 1 (10) gradient descent step and 3 (3) epochs.

\section{Distillation Results}

A comparison is drawn between distillation results of classical LeNet model and the proposed quantum LeNet model. For the quantum LeNet model in particular, we show results under, 1. no residual connection and the measurement observable of Pauli-Z unitary operator, 2. no residual connection and the measurement observable of a trainable Hermitian operator and 3. with a residual connection and the measurement observable of a trainable Hermitian operator. 

\textbf{Fixed initialization:} For both the datasets, the results are tabulated in Table \ref{tab:fixed_init_results}. We observe that for the quantum LeNet model, the variant having neither residual connection nor Hermitian observable performs the worst ($82.3\%$ MNIST, $28.5\%$ Cifar-10), followed by the variant having no residual connection but having Hermitian observable ($86.5\%$ MNIST, $31.3\%$ Cifar-10). The best performance is exhibited by the variant having both residual connections and Hermitian observable ($91.8\%$ MNIST, $50.3\%$ Cifar-10). These results indicate that having no residual connection and restrictive Pauli-Z unitary observable can cause issues like vanishing gradients which leads to relatively lower inferencing performance. Next, when the Pauli-Z operator is replaced with a trainable Hermitian operator, it does not necessarily preserve the norm of the quantum state and thus, the restrictive output bounds $[-1, 1]$ are removed which reduces issues like vanishing gradients. Finally, adding residual connection over the QNN layer further mitigates the vanishing gradients. We also show the final distilled images giving the aforementioned results in Fig. \ref{fig:distilled_images} for visual comparison. The plot shows results for both classical and quantum LeNet models and includes MNIST and Cifar-10 datasets. Similar to the results of the original work \cite{wang2018dataset}, the distilled images for the case of fixed initialization do not have visual resemblance to the original class images. For example, class 0 image for distilled MNIST images for both quantum and classical LeNets do not have the shape of `0' in it. Same can be said for rest of the classes of images and for Cifar-10 dataset as well.

\begin{table}[t]
\centering
\caption{Post-distillation inferencing accuracy of quantum LeNet having fixed initialization for cases having trainable and non-trainable Hermitian observable.}
\label{tab:herm_results}
\begin{tabular}{|l|ll|}
\hline
\multirow{2}{*}{\textbf{Dataset}} & \multicolumn{2}{l|}{\textbf{Hermitian}}                          \\ \cline{2-3} 
                                  & \multicolumn{1}{l|}{\textbf{Trainable}} & \textbf{Non-trainable} \\ \hline
\textit{MNIST}                    & \multicolumn{1}{l|}{91.9\%}             & 90.1\%                 \\ \hline
\textit{Cifar-10}                 & \multicolumn{1}{l|}{50.3\%}             & 49.0\%                 \\ \hline
\end{tabular}
\vspace{-3mm}
\end{table}

\textbf{Trainable vs non-trainable Hermitian:} While having a trainable Hermitian can help in getting better inferencing performance, there is a chance that this trainability can make the loss landscape more complex and induce an instability in the distillation process. So, we consider another scenario where we initialize the Hermitian randomly and do not train it over the course of dataset distillation process and compare it with the case where the Hermitian is trainable. In this comparison, we retain the residual skip connection in both the cases to obtain the best distillation results (Table \ref{tab:herm_results}). For the trainable Hermitian, we borrow the results from Table \ref{tab:fixed_init_results} (R, H case). We note that removing trainability of the Hermitian minimally impacts the results. For MNIST (Cifar-10) dataset, we note an accuracy reduction of $1.8\%$ ($1.3\%$). 

\section{Discussion \& Limitations}
\textbf{Lower quantum LeNet performance compared to classical LeNet:} From the results, we note that the best variant of quantum LeNet performs slightly worse compared to classical LeNet. For example, in Table \ref{tab:fixed_init_results} we note that trainable Hermitian observable measurement in QNN layer along with residual connection added shows $91.6\%$ ($50.3\%$) post-distillation inferencing accuracy for quantum LeNet compared to $94\%$ ($54\%$) accuracy for classical LeNet and MNIST (Cifar-10) dataset. A potential reason for this is the usage of amplitude embedding in the QNN layer. While amplitude embedding is efficiently able to map $2^n$ features on to $n$ qubits ($64$ features on $6$ qubits in this case), it has been shown that the average of encoded quantum states created by amplitude embedding tend to concentrate towards a specific state \cite{wang2025limitations}. This concentration tends to create a loss barrier phenomenon for QNN where the loss has a lower bound below which it cannot be optimized. This can potentially be mitigated by further classically reducing $2^n$ features using additional dense layers to $n$ features and embedding the features in the QNN using angle embedding \cite{nielsen2010quantum}.

\textbf{Practical implementation of Hermitian observables:}
In general, a single qubit Hermitian observable can be decomposed as a linear combination of Pauli matrices and the Identity matrix \cite{nielsen2010quantum}. For real values $a_0, a_x, a_y,a_z \in \mathbb{R}$, a single qubit Hermitian can be represented as
$$
O = a_0 I + a_x X + a_y Y + a_z Z
$$
Since the individual Pauli and Identity observables are practically implementable on a quantum computer \cite{IBMQuantum2025, MicrosoftQuantum2025}, their linear combination which is the Hermitian $O$ is also implementable on a quantum computer. This is good enough for the quantum LeNet architecture, since the QNN layer only uses single qubit Hermitian observables. An extension to this however, would be to implement multi-qubit Hermitian observables where each multi-qubit Hermitian would be represented as linear combination of tensor products of Pauli matrices. Such an observable can be useful in capturing entanglement information between the features, however it can be expensive to implement them practically as an $n$ qubit Hermitian observable can take upto $4^n$ Pauli terms \cite{guo2023estimating}.

\textbf{Adding quantum part in feature extraction portion of LeNet:} One can add quantum components to the feature extraction portion of LeNet architecture. Quantum convolutional layers, also known as quanvolutional layers were introduced in \cite{henderson2020quanvolutional} where traditional convolutional layers were replaced by random quantum circuits which served as the core transformational layers. The Quanvolutional Neural Network outperformed the CNN by roughly $5\%$ in terms of inferencing accuracy for MNIST dataset compared to classical Convolutional Neural Network (CNN). By replacing some convolution layers with quanvolution layers, we can potentially see increase in post-distillation inferencing performance on MNIST (and even Cifar-10) dataset(s).


\section{Conclusion}
In this work, we proposed a novel quantum LeNet model and used it for performing dataset distillation process. Our results show that the best quantum variant having trainable Hermitian observable and residual connection added performs close to the classical LeNet model in terms of inferencing accuracy for both MNIST and Cifar-10 datasets.

\section*{Acknowledgements}
We acknowledge the usage of Pennylane for performing all the experiments. This work is supported in parts by NSF (CNS-2129675 and CCF-2210963) and Intel’s gift.

\bibliographystyle{IEEEtran}
\bibliography{references}

\begin{thebibliography}{10}
\providecommand{\url}[1]{#1}
\csname url@samestyle\endcsname
\providecommand{\newblock}{\relax}
\providecommand{\bibinfo}[2]{#2}
\providecommand{\BIBentrySTDinterwordspacing}{\spaceskip=0pt\relax}
\providecommand{\BIBentryALTinterwordstretchfactor}{4}
\providecommand{\BIBentryALTinterwordspacing}{\spaceskip=\fontdimen2\font plus
\BIBentryALTinterwordstretchfactor\fontdimen3\font minus \fontdimen4\font\relax}
\providecommand{\BIBforeignlanguage}[2]{{%
\expandafter\ifx\csname l@#1\endcsname\relax
\typeout{** WARNING: IEEEtran.bst: No hyphenation pattern has been}%
\typeout{** loaded for the language `#1'. Using the pattern for}%
\typeout{** the default language instead.}%
\else
\language=\csname l@#1\endcsname
\fi
#2}}
\providecommand{\BIBdecl}{\relax}
\BIBdecl

\bibitem{abbas2021power}
A.~Abbas, D.~Sutter, C.~Zoufal, A.~Lucchi, A.~Figalli, and S.~Woerner, ``The power of quantum neural networks,'' \emph{Nature Computational Science}, vol.~1, no.~6, pp. 403--409, 2021.

\bibitem{stein2023benchmarking}
J.~Stein, M.~Poppel, P.~Adamczyk, R.~Fabry, Z.~Wu, M.~K{\"o}lle, J.~N{\"u}{\ss}lein, D.~Schuman, P.~Altmann, T.~Ehmer \emph{et~al.}, ``Benchmarking quantum surrogate models on scarce and noisy data,'' \emph{arXiv preprint arXiv:2306.05042}, 2023.

\bibitem{kashif2024computational}
M.~Kashif, A.~Marchisio, and M.~Shafique, ``Computational advantage in hybrid quantum neural networks: Myth or reality?'' \emph{arXiv preprint arXiv:2412.04991}, 2024.

\bibitem{Graps2024}
\BIBentryALTinterwordspacing
A.~Graps. (2024, 10) Quantum for {AI} costs from a diagnostic benchmark. Accessed: March 5, 2025. [Online]. Available: \url{https://quantumcomputingreport.com/quantum-for-ai-costs-from-a-diagnostic-benchmark/}
\BIBentrySTDinterwordspacing

\bibitem{kordzanganeh2023benchmarking}
M.~Kordzanganeh, M.~Buchberger, B.~Kyriacou, M.~Povolotskii, W.~Fischer, A.~Kurkin, W.~Somogyi, A.~Sagingalieva, M.~Pflitsch, and A.~Melnikov, ``Benchmarking simulated and physical quantum processing units using quantum and hybrid algorithms,'' \emph{Advanced Quantum Technologies}, vol.~6, no.~8, p. 2300043, 2023.

\bibitem{wang2018dataset}
T.~Wang, J.-Y. Zhu, A.~Torralba, and A.~A. Efros, ``Dataset distillation,'' \emph{arXiv preprint arXiv:1811.10959}, 2018.

\bibitem{lecun1998gradient}
Y.~LeCun, L.~Bottou, Y.~Bengio, and P.~Haffner, ``Gradient-based learning applied to document recognition,'' \emph{Proceedings of the IEEE}, vol.~86, no.~11, pp. 2278--2324, 1998.

\bibitem{schuld2020circuit}
M.~Schuld, A.~Bocharov, K.~M. Svore, and N.~Wiebe, ``Circuit-centric quantum classifiers,'' \emph{Physical Review A}, vol. 101, no.~3, p. 032308, 2020.

\bibitem{kanamori2020quantum}
Y.~Kanamori and S.-M. Yoo, ``Quantum computing: principles and applications,'' \emph{Journal of International Technology and Information Management}, vol.~29, no.~2, pp. 43--71, 2020.

\bibitem{nielsen2010quantum}
M.~A. Nielsen and I.~L. Chuang, \emph{Quantum computation and quantum information}.\hskip 1em plus 0.5em minus 0.4em\relax Cambridge university press, 2010.

\bibitem{li2021survey}
Z.~Li, F.~Liu, W.~Yang, S.~Peng, and J.~Zhou, ``A survey of convolutional neural networks: analysis, applications, and prospects,'' \emph{IEEE transactions on neural networks and learning systems}, vol.~33, no.~12, pp. 6999--7019, 2021.

\bibitem{cazenavette2022dataset}
G.~Cazenavette, T.~Wang, A.~Torralba, A.~A. Efros, and J.-Y. Zhu, ``Dataset distillation by matching training trajectories,'' in \emph{Proceedings of the IEEE/CVF Conference on Computer Vision and Pattern Recognition}, 2022, pp. 4750--4759.

\bibitem{zhao2020dataset}
B.~Zhao, K.~R. Mopuri, and H.~Bilen, ``Dataset condensation with gradient matching,'' \emph{arXiv preprint arXiv:2006.05929}, 2020.

\bibitem{zhao2023dataset}
B.~Zhao and H.~Bilen, ``Dataset condensation with distribution matching,'' in \emph{Proceedings of the IEEE/CVF Winter Conference on Applications of Computer Vision}, 2023, pp. 6514--6523.

\bibitem{letcher2024tight}
A.~Letcher, S.~Woerner, and C.~Zoufal, ``Tight and efficient gradient bounds for parameterized quantum circuits,'' \emph{Quantum}, vol.~8, p. 1484, 2024.

\bibitem{he2016deep}
K.~He, X.~Zhang, S.~Ren, and J.~Sun, ``Deep residual learning for image recognition,'' in \emph{Proceedings of the IEEE conference on computer vision and pattern recognition}, 2016, pp. 770--778.

\bibitem{wang2025limitations}
X.~Wang, Y.~Wang, B.~Qi, and R.~Wu, ``Limitations of amplitude encoding on quantum classification,'' \emph{arXiv preprint arXiv:2503.01545}, 2025.

\bibitem{IBMQuantum2025}
\BIBentryALTinterwordspacing
{IBM Quantum}, ``Specify observables in the pauli basis,'' IBM Quantum Documentation, 2025, accessed: March 12, 2025. [Online]. Available: \url{https://docs.quantum.ibm.com/guides/specify-observables-pauli}
\BIBentrySTDinterwordspacing

\bibitem{MicrosoftQuantum2025}
\BIBentryALTinterwordspacing
{Microsoft}, ``Single-qubit and multi-qubit {Pauli} measurements,'' Microsoft Azure Quantum Documentation, 2025, accessed: March 12, 2025. [Online]. Available: \url{https://learn.microsoft.com/en-us/azure/quantum/concepts-pauli-measurements}
\BIBentrySTDinterwordspacing

\bibitem{guo2023estimating}
N.~Guo, F.~Pan, and P.~Rebentrost, ``Estimating properties of a quantum state by importance-sampled operator shadows,'' \emph{arXiv preprint arXiv:2305.09374}, 2023.

\bibitem{henderson2020quanvolutional}
M.~Henderson, S.~Shakya, S.~Pradhan, and T.~Cook, ``Quanvolutional neural networks: powering image recognition with quantum circuits,'' \emph{Quantum Machine Intelligence}, vol.~2, no.~1, p.~2, 2020.

\end{thebibliography}

\end{document}